\documentclass[letterpaper, 10 pt, conference]{ieeeconf}  

\IEEEoverridecommandlockouts                              

\title{\LARGE \bf
Semantic Image Based Geolocation Given a Map
}

\author{Arsalan Mousavian and Jana Ko\v{s}eck\'a
\thanks{Arsalan Mousavian and Jana Ko\v{s}eck\'a are with the Computer Science Department, Volgenau School of Engineering at George Mason University, Fairfax, VA 22030, USA. {\tt\small amousavi@gmu.edu, kosecka@gmu.edu}.}}\vspace{3mm}

\usepackage{times}
\usepackage{epsfig}
\usepackage{graphicx}
\usepackage{amsmath}
\usepackage{amssymb}
\usepackage{dsfont}

\begin{document}
\maketitle
\thispagestyle{empty}
\pagestyle{empty}
\begin{abstract}
The problem visual place recognition is commonly used strategy for localization.
Most successful appearance based methods typically rely on a large database of views endowed with local or global image descriptors and strive to retrieve the views of the same location. 
The quality of the results is often affected by the density of the reference views and the robustness of the image representation with respect to viewpoint variations, clutter and seasonal changes.  
In this work we present an approach for geo-locating a novel view and determining camera location and orientation  using a map and a sparse set of geo-tagged reference views. We propose a novel technique for detection and identification of building facades from geo-tagged reference view using the map and geometry of the building facades. We compute the likelihood of camera location and orientation of the query images using the detected landmark (building) identities from reference views, 2D map of the environment, and geometry of building facades.
We evaluate our approach for building identification and geo-localization on a new challenging outdoors urban dataset exhibiting large variations in appearance and viewpoint.
\end{abstract}

\section{Introduction}
\begin{figure}[t]
\includegraphics[width=0.48\textwidth]{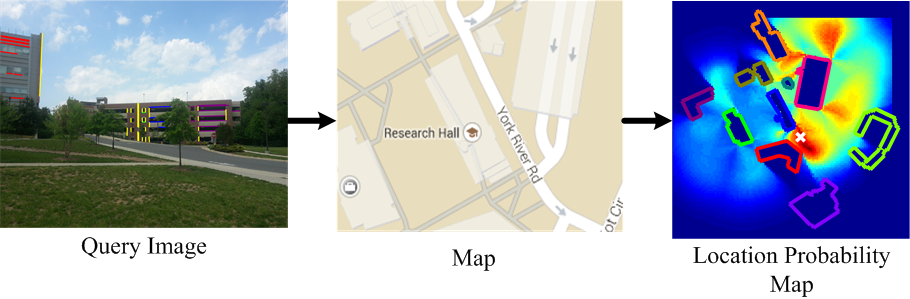}
\caption{Overview of the approach: for each query image we identify which buildings are in the image as well as the orientation of building facades. We then use the identity of the buildings and orientation of building facades and the map, to find the probability distribution for the location of image.}
\label{fig:ICCV}
\end{figure}

The problem of image based geolocation is a problem of determining the geographic location of a query view.  The geolocation takes the location recognition one step further and attempts to estimate the exact geo-pose of the query view, which can be used in the context of various localization strategies. 
The approaches considered in the past, explore variety of visual and non-visual cues and use single view, video sequences or large geotagged reference datasets to tackle the problem.  The techniques which use large datasets of  geo-tagged views focus on the scalability aspects~\cite{Hays-CVPR08}  with the goal of approximately locating the query views. Alternative methods use densely sampled images of smaller geographical area and approach the problem by matching local features followed by geometric verification and triangulation~\cite{ZamirECCV10}. This requires retrieval of at least two images from the reference set with sufficient overlap for triangulation.
In this work, we explore the setting where the reference images come from smaller geo-graphical area and are sparsely sampled having small or no overlap. Consequently, it is not always possible to retrieve two images to enable triangulation of the position. In this work we leverage the meta-data of publicly available maps, such as {\tt OpenStreetMap}, that has GPS coordinates of building corners. We use a sparse set of views to guide the approximate location recognition, followed by the estimation of 3D geometry from the single query view. The obtained 3D model along with the building identity obtained from appearance based location recognition is then used to estimate the pose likelihood of the image given the map of the area.  The computation of the pose likelihood yields the probability distribution for the possible locations of the query image.
The ingredients of the approach are shown in Figure~\ref{fig:ICCV}. 

\begin{figure}
\begin{tabular}{c@{\hspace{1mm}}c@{\hspace{1mm}}c}
\includegraphics[width=0.15\textwidth , height=0.15\textwidth]{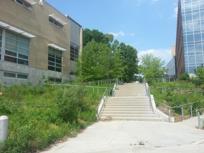}&
\includegraphics[width=0.15\textwidth]{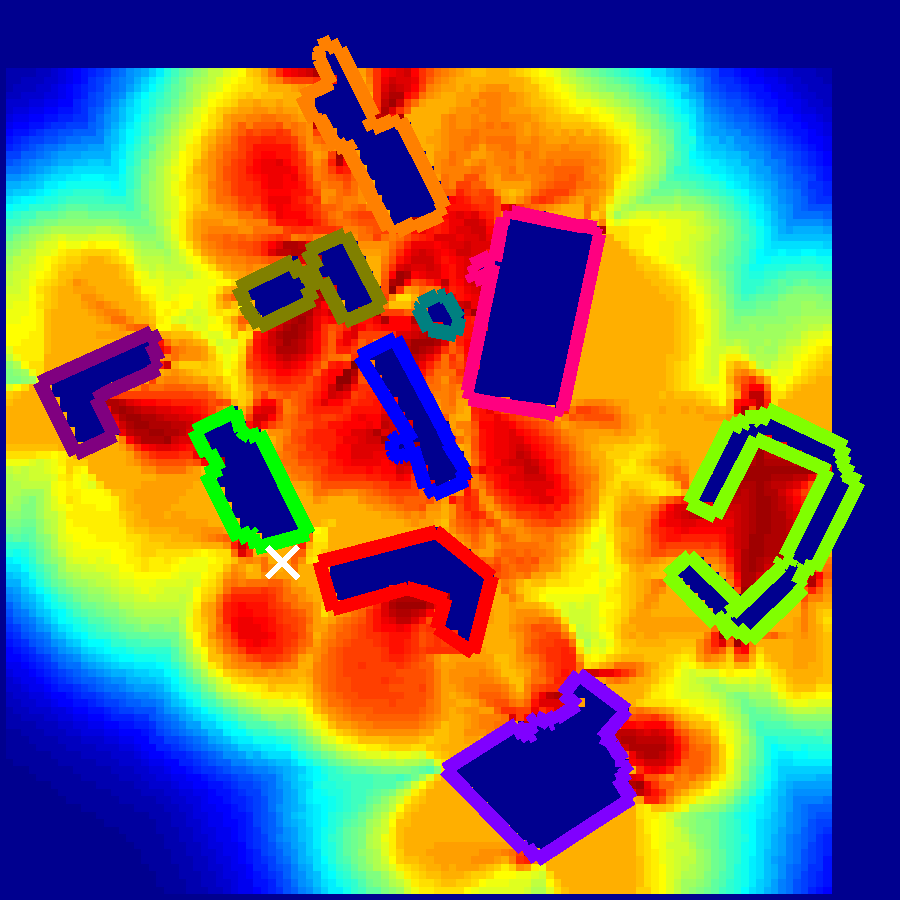}&
\includegraphics[width=0.15\textwidth]{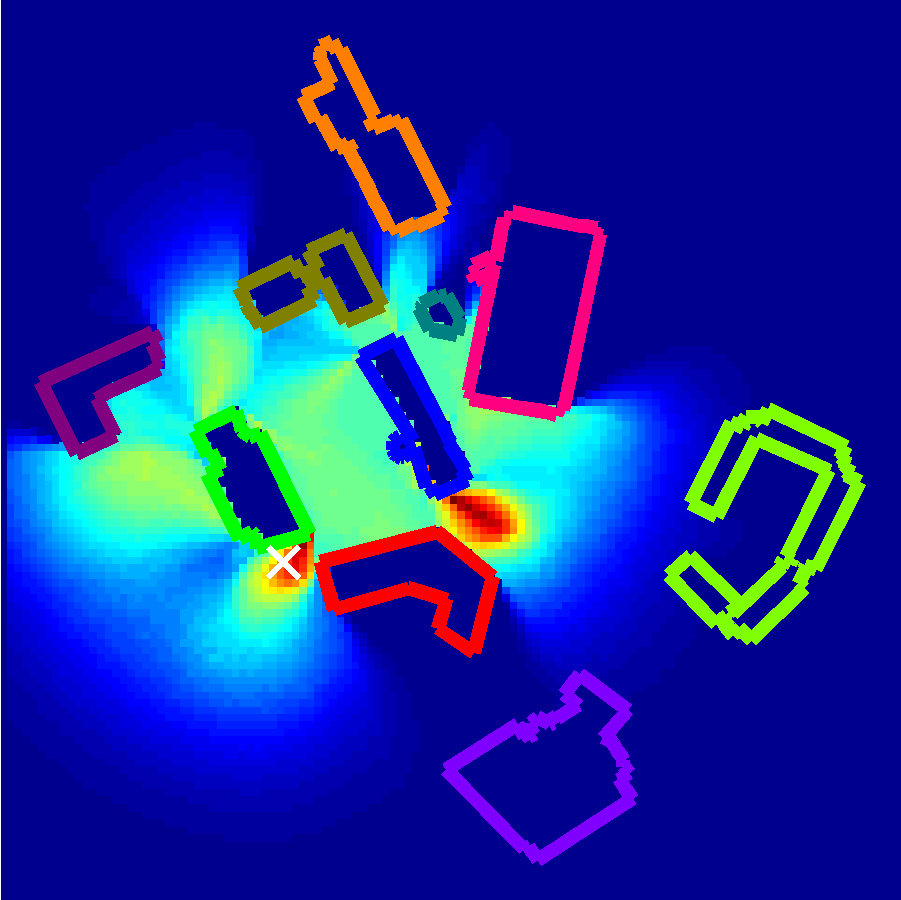}\\
\includegraphics[width=0.15\textwidth , height=0.15\textwidth]{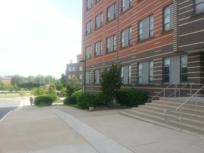}&
\includegraphics[width=0.15\textwidth]{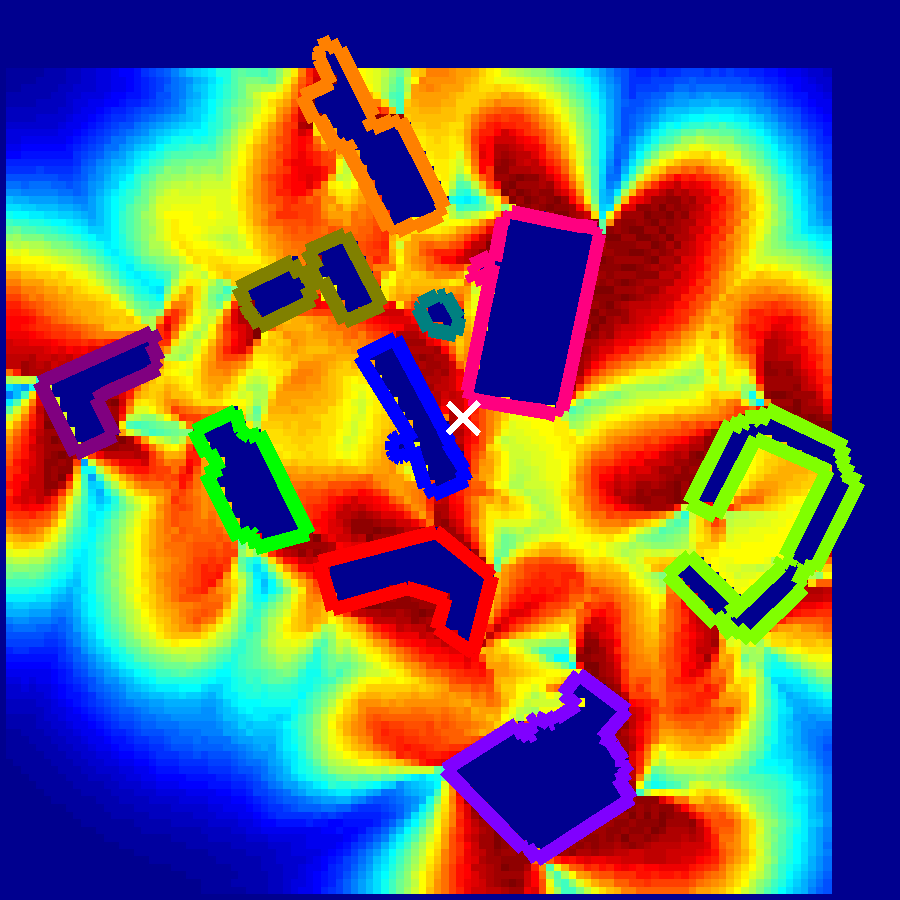}&
\includegraphics[width=0.15\textwidth]{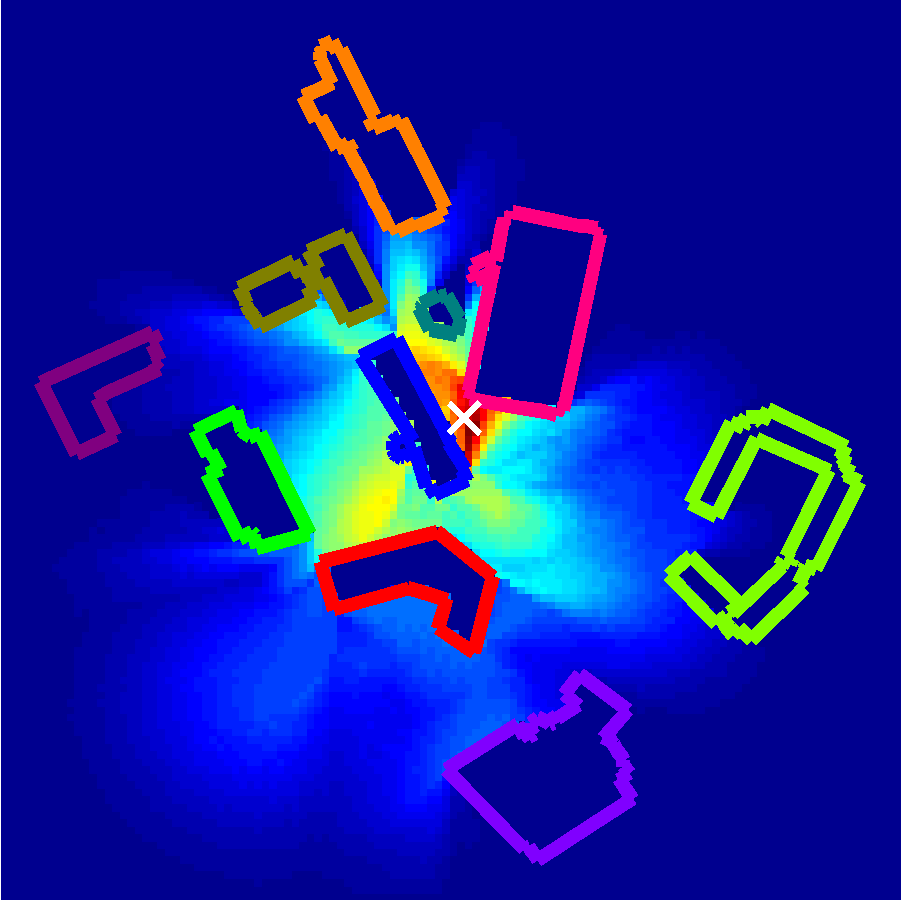}\\
\includegraphics[width=0.15\textwidth , height=0.15\textwidth]{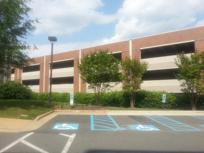}&
\includegraphics[width=0.15\textwidth]{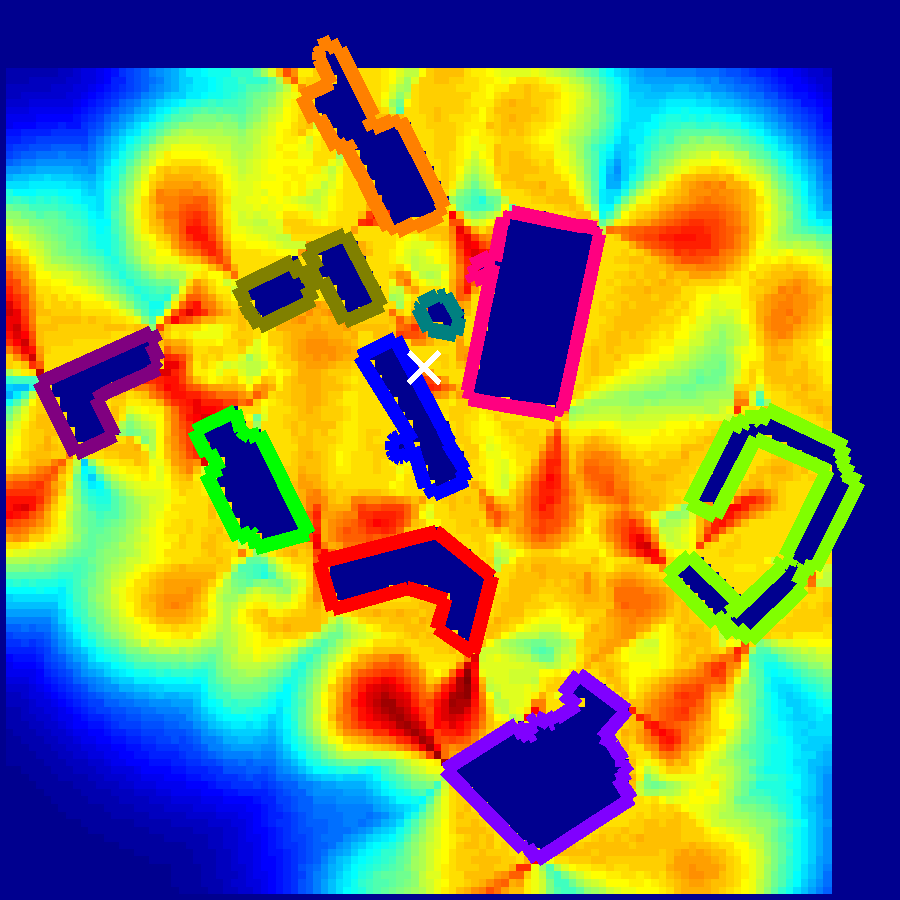}&
\includegraphics[width=0.15\textwidth]{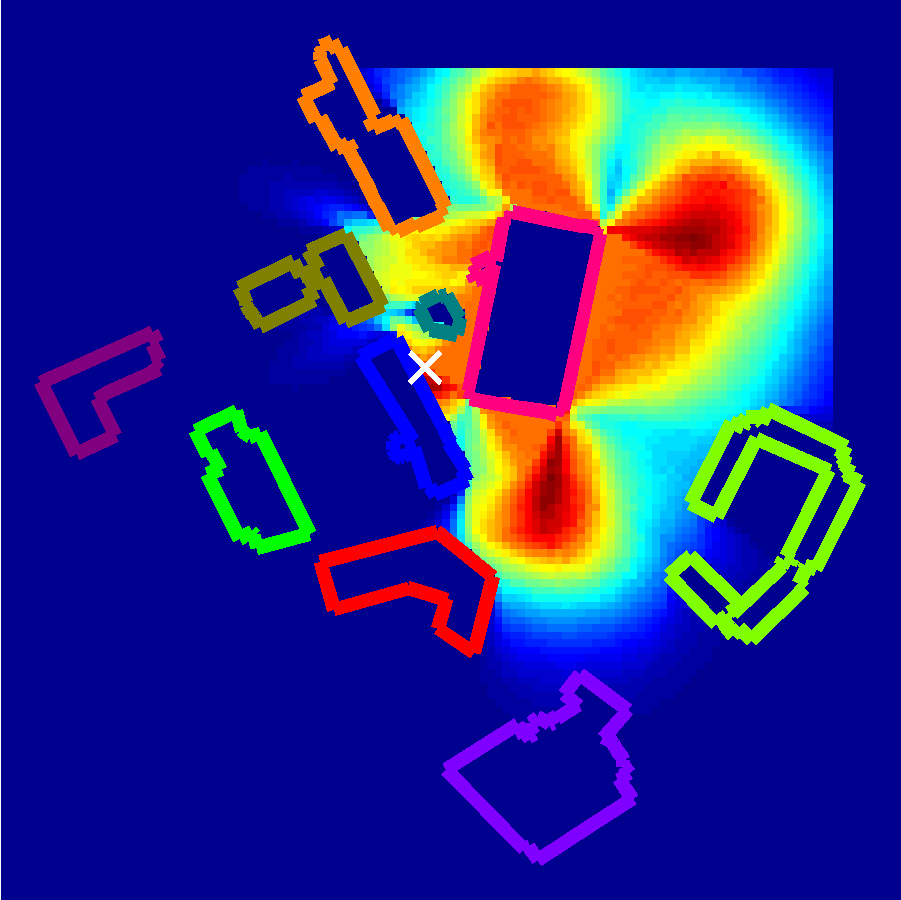}\\
\end{tabular}
\caption{Comparison of localization with building identification on top of the geometric signature of the image. Left: Query image. Middle: Probability map for location of the image using Geometric Signature Right: Probability map for location of the image using Geometric Signature + Building Facade Identity. The white cross in the probability maps shows the ground truth location of the image (Best viewed in electronic version).}
\label{fig:FOV}
\end{figure}

The idea of using the 2D map of the environment and the geometry of building facade for localization and mapping has been explored before. In \cite{BURGARD-SPRINGER-2011}, aerial images are used to add more global constraints within the graph-representation of the SLAM to reduce the error in the final maps.  Authors in \cite{Cham-CVPR-2010} and \cite{David-IROS-2011} proposed to localize  the $360^\circ$ panorama images by 
matching \emph{"geometric signature"} of the image to the 2D map of the environment. Geometric signature of the image is the representation for all of the visible building facades in the image and their relative orientation with respect to each other. The larger the field of view, the more building facades are visible and yielding more unique the geometric signature. However, the performance degrades dramatically if the same approach is applied to a regular standard FOV views. To the best of our knowledge, we are the first to address the problem of geo-localization of non-panoramic images given a 2D map and a sparse set of reference views.  Figure~\ref{fig:FOV} shows the computed location probability map for few query images using only 3D geometric signature  (middle column) and using both 3D geometry and appearance (right column) of the query view.
Note that using only the geometric signature with the limited FOV images produces significantly more ambiguous location probability map. 

\paragraph{Outline} 
In the first stage of the approach we compute the camera orientation of a sparse set of geotagged reference views, which enables us to identify and segment the buildings with the aid of the map. For the geolocation of a query view, first the location recognition and building identification is done by matching the query view to a reference set, followed by estimation of 3D building facades. The whole framework is formulated in a probabilistic setting where the evidence from  the appearance based matching, building identification and 3D geometry is combined to compute the came pose likelihood given a map.    

\section{Related Works}
The problem of geolocation has been studied extensively by many and the existing approaches vary in the proposed image representation and associated matching strategy as well as datasets used for evaluation.  
One line of work, which is the prerequisite for geolocation using large reference sets 
is the retrieval of nearest views.  Here the considered baseline method is often the bag-of-visual-words representation of images, 
followed by spatial verification of the top retrieved images using geometric constraints~\cite{Philbin-CVPR07}. The  improvements of this methods include learning better vocabularies, developing better quantization and spatial verification methods~\cite{ChumECCV10,ZamirECCV10,Zhang-3DPVT06}.  In contrast to image based retrieval, in visual place recognition, there is often additional structural information available which can be exploited towards the task. 
For example authors in~\cite{SnavelyCVPR13} built adjacency matrix between reference images and then find the clusters within training images. During the recall, they first classify the image by assigning it to most likely image cluster and then retrieve the image using TF-IDF weighting scheme. In~\cite{Knopp-ECCV10}, authors learn the confusion weight for each feature exploiting the geographic location of the unweighted retrieved views and in~\cite{GronatCVPR13} they train exemplar SVM per image in the training set where the negative examples are the images which have similar cosine distance but they are far away.  
The geolocation takes the location recognition one step further and attempts to estimate the exact geo-pose of the query view. In order to be able to estimate exact pose, it is necessary that at least two overlapping views are found, so the pose of the query view can be triangulated~\cite{ZamirECCV10}.  Techniques which relied on 3D reconstruction, matched 3D models to either 3D point clouds~\cite{Snavely-ECCV12} or maps~\cite{RadekICCV09}. 
Another line of work was proposed in \cite{McManusRSS2014} to find discriminative and robust patches at each location from reference images of that location taken at different time of  year in the training time. At the test time, they localized the query images by detecting the learned patches in the query view and assigning the query location to the location of corresponding patches from training. Alternative approaches were also used by \cite{RatSLAMTRO08} to overcome the challenges of viewpoint variation by using low resolution features for each location.

The technique proposed here exploits 3D geometric information and is related to different efforts towards single view reconstruction of man-made structures~\cite{Han-PAMI-2009,Hoiem-ICCV-2005}. In the presented approach we not only rely on reliable techniques for reconstruction of local planar coordinate frames, but also  combine geometric techniques with the semantic segmentation. 
The use of geometric signatures of the image-based for localization has been done before for panoramic images.  Cham et al \cite{Cham-CVPR-2010}, found the depth ratio of the boundary lines of each planar facade and look for positions in the map where the observed depth ratio of planes can be seen using geometric hashing.
Unlike our method, their method was susceptible to occlusion of boundaries between building facades. David and Ho \cite{David-IROS-2011}, proposed a different geometric signature for each panoramic image. They defined geometric signature by casting rays from each position at different directions and check whether it hits any building in the map or not. One of the downside of their method is that if the observation point is far from subset of the buildings, they cannot detect any lines belonging to building facades but they are taken into account when computing the signature from the map. In our geometric signature, the importance of building facades which are far away is diminished to account for the fact that they can be occluded by other buildings or too far to be detected reliably.  Figure~\ref{fig:FOV} (middle column) shows that if we rely only on geometric signature for the localization of non-panoramic images the probability map for location of the image becomes more ambiguous.


While the current work focuses on a static setting, the development of the proper pose likelihood model  is the main ingredient of robotic localization techniques given a map, which typically rely on geometric information. Semantic localization in~\cite{AtanasovRSS14} uses traditional object detection pipelines~\cite{DalalTriggsCVPR05}  to generate hypotheses about presence of objects and their extent and bearing in images, followed by a particle filter based localization.  The current work naturally extends the type of semantic information which can be considered for semantic localization tasks and demonstrates how to use it in the settings for which maps are readily available.

\section{Proposed Method}
\begin{figure}[t]
  \begin{tabular}{l} 
   \includegraphics[width=0.48\linewidth]{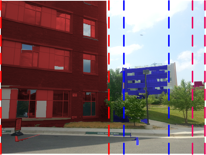}\hspace{1mm}\includegraphics[width=0.48\linewidth,height=3cm]{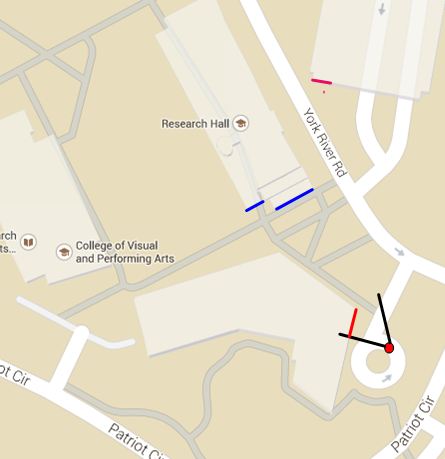}  \\
   \includegraphics[width=0.48\linewidth]{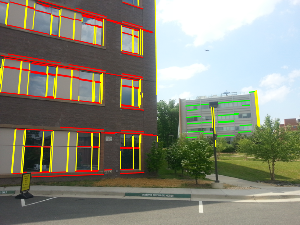}\,
      \includegraphics[width=0.48\linewidth]{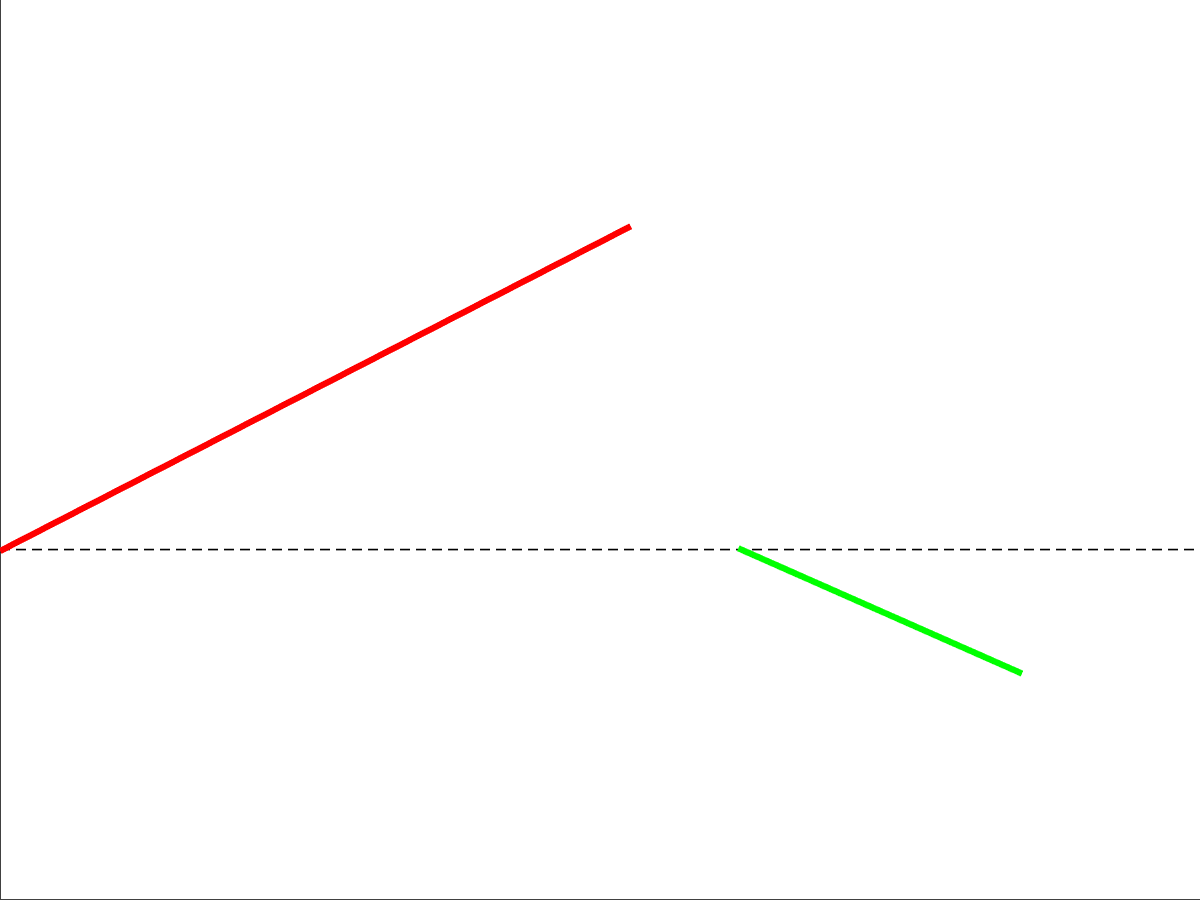}\,
 \end{tabular}
 \caption{Top: Inferred building segmentation and identification for an unlabeled geo-tagged image (different colors indicate different building identities) and the corresponding map with geometric signature where the estimated camera pose in addition to visible facades in the map are illustrated with the corresponding colors.
 Bottom: line segments used for plane normal estimation; plane normal orientation estimates. yellow lines represent vertical lines and horizontal lines on each planar building facade represent with different color. The orientations are shown with corresponding color.}  
 \label{fig:overview}
\end{figure}

We are given a geographical area sparsely populated by geo-tagged views along with the 2D map of the environment  
containing the coordinates of building corners and  building identities.
The goal is to compute the location and camera orientation of a novel query view. 
We can partition the approach into two tasks:
\begin{enumerate}
\item the task of estimation of camera orientation for a  set of geo-tagged reference views and
automated building identification and segmentation given the map; 
 \item the task of computing the geolocation of a novel query view, given the map and pre-processed reference set of geo-tagged views. 
\end{enumerate}

\subsection{Reference Set Building Identification}
Given a reference set of images geo-tagged with longitude and latitude, the goal of the first stage is to determine the camera orientation
and building identities in the reference views with the aid of the map. Instead of tedious manual labelling and segmentation of the buildings in the reference images, we first estimate the camera orientation and then project the map to individual views, assigning building identities to image regions. Due to commonly encountered errors in GPS coordinates, we sample possible orientations and nearby locations and  evaluate the likelihood of each pose with respect to the map containing geometric information about polygonal boundaries of the buildings along with their identity. To compute the camera pose likelihood of the reference view,  we detect buildings in images and recover the 3D facade normals  from the vanishing points.  At last we compute probabilities of building identities for image regions labelled as buildings in the image by semantic segmentation algorithm. Figure~\ref{fig:overview} summarizes the approach with the details provided next.

\paragraph{Decomposition of Building Facades}

\begin{figure}
 \begin{tabular}{c@{\hspace{1mm}}c}
    
      \includegraphics[width=0.48\linewidth]{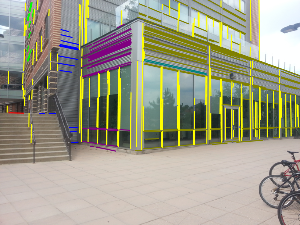} &
      \includegraphics[width=0.48\linewidth]{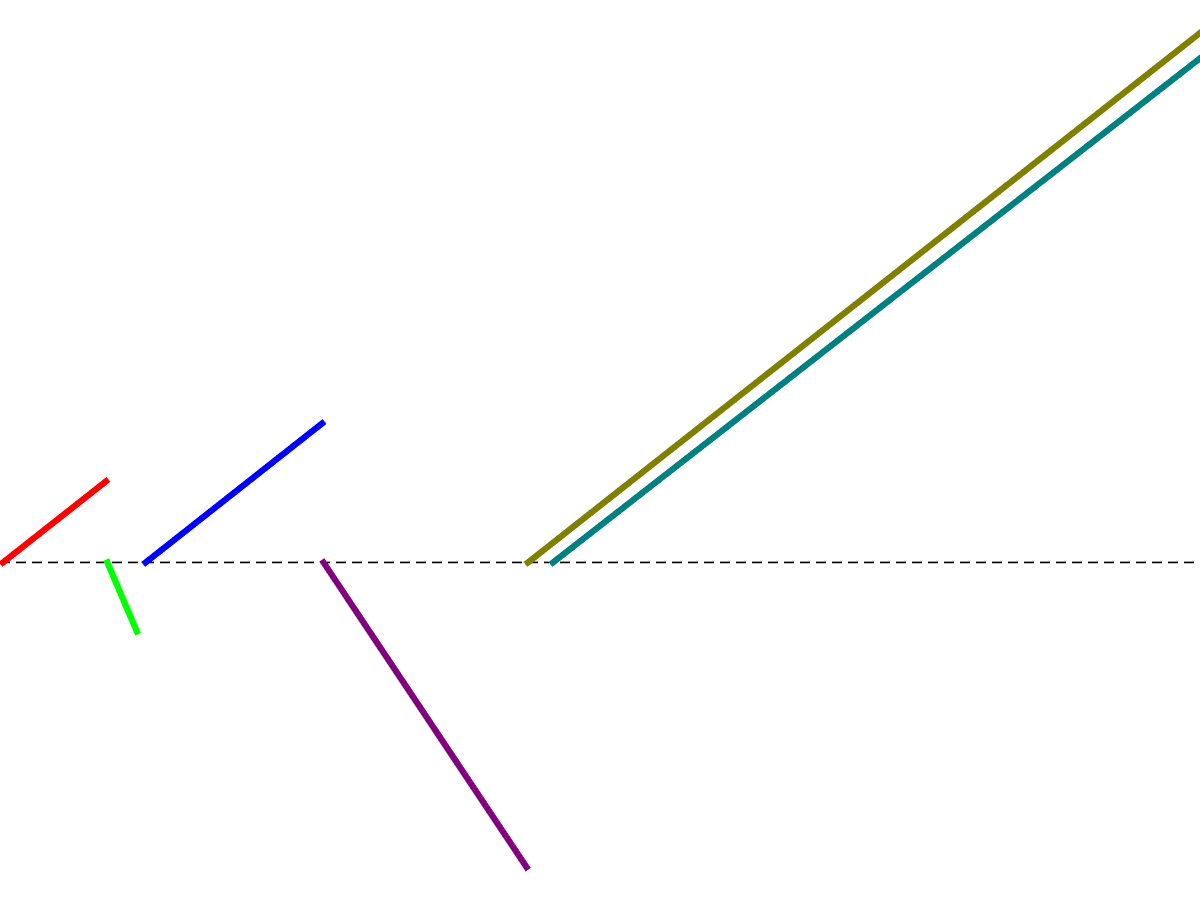}\\
      \includegraphics[width=0.48\linewidth]{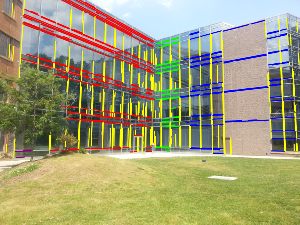}&
      \includegraphics[width=0.48\linewidth]{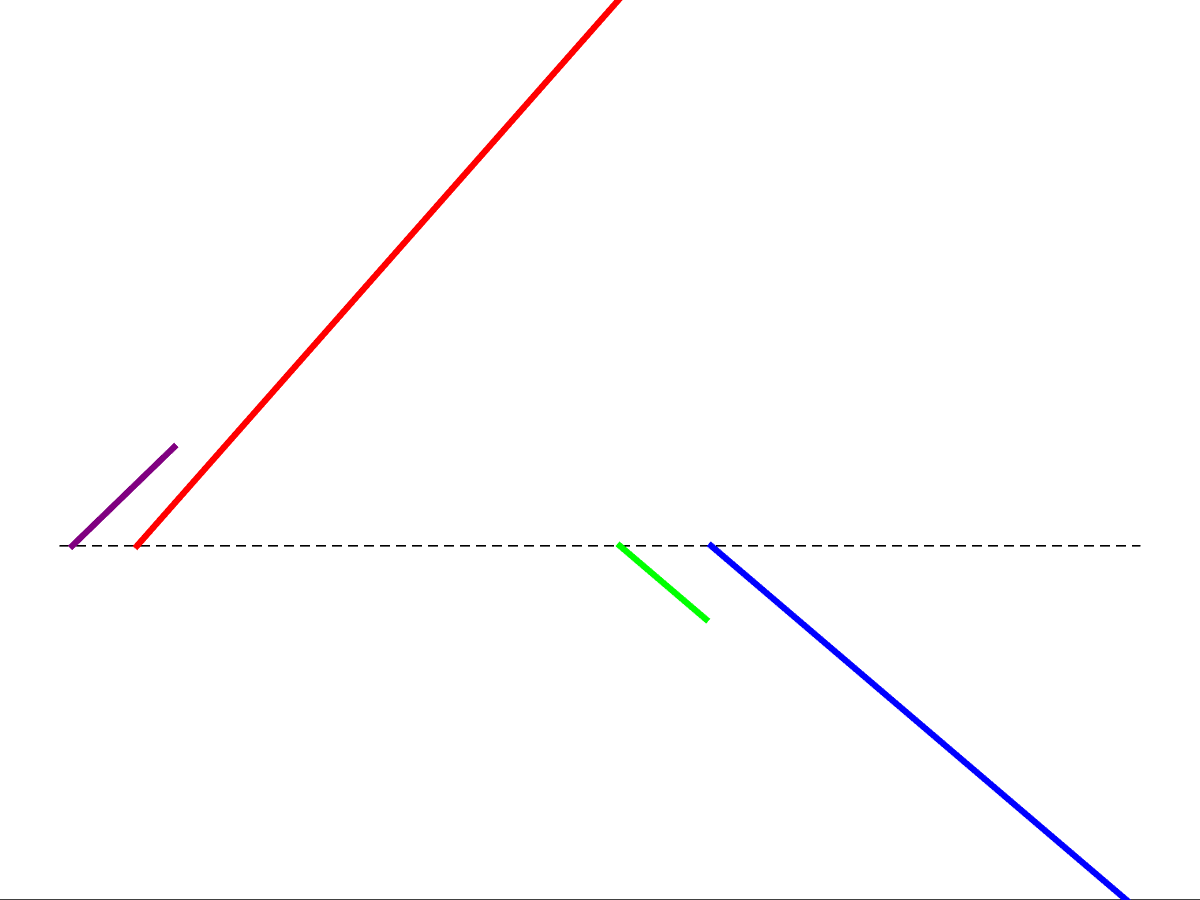}\\
      Detected Lines & Plane Orientation


 \end{tabular}
 \caption{Illustration of Plane Orientation Estimation. In the detected lines column each group of lines which are associated with the same vanishing point are shown in the same color. The plane orientation columns show the orientation for each of the planes we detect from bird eye view. The color of each plane orientation line corresponds to the color of horizontal lines in the image. The dashed line represent the orientation of the camera's principal plane. Note that each building side might be decomposed to multiple overlapping planes as it is shown in some of the images.}
 \label{fig:PlaneOrientation}
 \end{figure}

 In order to decompose each image $\mathcal{I}$ to a set of planes which belonging to building facades, we use the method of \cite{Tretyak-IJCV-2012} to find the line segments in the image and their assignment to vanishing points (VP). The detected line segments can be also on the background (e.g. tree). We also compute a semantic segmentation of the image using a methods similar~ \cite{Tighe-ICCV11}, where regions in the image are automatically labelled as one of the 5 semantic categories {\em building, tree, road, sky, car}.  More details about this stage can be found in next sections. To eliminate the unwanted  line segments, we reject the lines which do not have any overlap with the areas classified as {\em building}. For each connected {\em building} region, we estimate local coordinate frames by determining the orientation of the planar facades. Each planar hypothesis $p_i \in P(\mathcal{I})$ in the image $\mathcal{I}$ is defined by its extent $[x_{start}^i,x_{end}^i]$ in the horizontal field of view and $\theta^i$ the orientation of the plane with respect to vertical y-axis. We call $x=x_{start}^i$ and $x=x_{end}^i$ cutting lines. An example of cutting lines is shown in Figure~\ref{fig:overview} with the dashed lines. Since there is no significant camera roll at the time of acquisition, the representation of facade extent by vertical lines is satisfactory. Now the problem of piecewise planar facade decomposition becomes a problem of finding the set of cutting lines defining the extent of each $p_i$ along with their orientation. Let $x=\{ \lambda_i: \lambda_i \in \Lambda \}$ be the set of cutting lines defining the extents of all the planes $p_i$ such that $p_i= [ x_{start}^i=\lambda_i,x_{end}^i=\lambda_{i+1},\theta^i ]$. In order to find $\lambda_i$, we use a greedy algorithm. For each $x$ coordinate, let $v(x)$ be the index of the vanishing point which has the most support from the line segments intersecting the vertical lines at $x$.   Vertical line at $x$ is a cutting line if the vanishing point label switches, i.e. $v(x) \neq v(x-1)$. Having decomposed each region labelled as {\em building} by semantic segmentation to a set of planes, we estimate the rotation matrix $R$ of each plane from the vanishing directions $l(x_{start}^i)$ and assuming planar motion obtain the relative angle $\theta^i$  with respect to $y$ axis of the camera frame.  Figure \ref{fig:overview} and \ref{fig:PlaneOrientation} show examples of line segment detection, plane decomposition and facades orientation estimation. Note that the method does not assume a single global Manhattan coordinate frame as many vanishing points are estimated. While at the moment the 3D reconstruction is limited to 3D plane normal estimation, the availability of the map can be also used effectively for estimating the distance of the planar facades from the cameras.
 
 \paragraph{Camera Pose Likelihood}

\begin{figure}
 \includegraphics[width=0.49\linewidth]{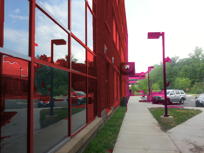}
 \includegraphics[width=0.49\linewidth]{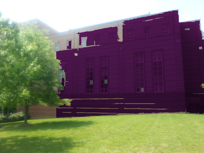}
 \caption{Decomposition of Buildings into faccades; each color coded region represents building identity, along with the faccade extent and
 plane orientation.  }. 
 \label{fig:faccades}
\end{figure}

In this section, we design the likelihood $p(\mathcal{I}|g,m)$ of an image $\mathcal{I}$ given a camera pose $g=(R,T)$ and the map. 
In the context of this problem we focus only on planar localization with $g = (x,y,\gamma)$, where 
$\gamma$ is the yaw angle, representing a rotation around vertical $y$ axis of the dominant Manhattan coordinate frame in the image. 
The goal of the camera pose likelihood is to quantify the  discrepancies between the information extracted from the image and the projection of the environment model (the map $m$), into the field of view of the virtual camera.  The image $\mathcal{I}$ model is comprised of set of $l$ planar facade hypotheses, their orientation and extent in the horizontal field of view of the image $\widehat{Z} = \{ \widehat{z}_i = [\widehat{x}_{start}^i,\widehat{x}_{end}^i, \widehat{\theta}_i] \}_{i=1}^l$  described in the previous section.  The quantities estimated from an image will be denoted by $\mbox{ } \widehat{ } \mbox{ }$.  We want the camera pose likelihood to account for the errors in camera orientations (hence change of viewing direction), errors in the orientation estimates of the planar facades, errors in detection of facades and estimates of their extent as well as discrepancies between the map and the image\footnote{Notice for example that a part of the building is occluded by vegetation, which is not represented in the map hence decreasing our confidence about the estimated extent of the building given the map}. 
Some examples of estimated facade hypotheses are in Figure~\ref{fig:faccades}. To simplify the notation we omit the map $m$ assuming that it is given. 
Given the camera position $(x,y)$  and the knowledge of the FOV of the camera, we can for each hypothesized orientation $\gamma$, compute the set of building facades from the map and their extents visible in the virtual view and their orientation. Denote the set of projected facade extents coordinates $Z = \{ z_i = [x_{start}^i, x_{end}^i, \theta^i, b_i, d^i]\}_{i=1}^m$.  We also record the shortest distance $d^i$ of the two facade endpoints from origin of the camera coordinate system and the building identity $b_i$.  This is a variation of an inverse sensor model which is commonly used in the robot localization literature given a map.

The likelihood will now be related to the similarity between the predicted set of observations $Z$ from the map $m$ and the estimates 
$\widehat{Z}$ obtained from the image $\mathcal{I}$.  The similarity between $Z$ and $\widehat{Z}$ is calculated by integrating (summing up) the evidence from all columns across the horizontal FOV.  Prior to computing the similarity we obtain a per column estimates of detected building facades obtained as average of the estimates per each connected component of the semantic segmentation in that column.  
The similarity measure between detected planar facades and their compatibility with the predicted models from the map is:  
\[ S(Z, \widehat{Z}) = \sum_{i=1}^m \sum_{j=1}^l \sum_{k=1}^{cols}  \mathds{1}(z_i^{k} \cap \widehat{z_j}^k \neq \emptyset) \times |  \cos(\theta^i-\widehat{\theta}^j) |  \times  w_i \]
where $w_i$ is the gaussian weight of the distance $d^i$ according to $\mathcal{N}(0, \sigma)$. 
This weight models the effect that buildings at further distances are typically harder to detect in images and  should be penalized accordingly. 
In order to be able to properly normalize the above compatibility measure we also need an expression for the maximum attainable 
similarity $S_{max}(Z, \widehat{Z})$. 
$S_{max}(Z, \widehat{Z})$ should have two  desirable properties: 1) it should assign maximum cosine distance to the columns  in which there is at least one projected plane (i.e. $z_k \neq \emptyset$),  2) it should quantify the contribution of columns where no planes are detected, but according to the map facades are projected in that part of FOV  (i.e. $z_i^k = \emptyset \wedge \widehat{z_j^k} \neq \emptyset$). The more different the pose is from the actual pose, the larger the second term gets and as a result decreases the overall likelihood.  These two characteristics are incorporated in the first and second term of Eq.\ref{eq:smax}   
\begin{equation}
\label{eq:smax}
S_{max}(Z, \widehat{Z}) = \sum_{i=1}^m \sum_{j=1}^l \sum_{k=1}^{cols} \mathds{1}(\widehat{z_j^k} \neq \emptyset)+ \nonumber\\ \mathds{1}(\widehat{z_j^k} = \emptyset \wedge {z_i^k} \neq \emptyset) \: w_i 
\end{equation}
Finally the likelihood $p(\mathcal{I}|g)$ is computed by:
\begin{equation}
\label{eq:LL}
p(\mathcal{I}|g) = \frac{S(Z, \widehat{Z}) }{S_{max}(Z, \widehat{Z})}
\end{equation}
Eq~\ref{eq:LL} can also be interpreted as modified intersection over union score. $S(Z,\widehat{Z})$ captures the similarity score in the intersecting part of $Z$ and $\widehat{Z}$. $S_{max}(Z, \widehat{Z})$ is the weighted union between the observations and the predictions from the map. 

The camera pose likelihood is computed for each reference view, sampling all possible orientations in increments of $3^\circ$. 
For certain locations this likelihood function has a single distinguished peak, in other settings it may have more local extrema. 

\paragraph{Assigning Building Identity to Pixels}
\begin{figure}[t]
 \begin{tabular}{c@{\hspace{1mm}}c@{\hspace{1mm}}c@{\hspace{1mm}}c}
     
\includegraphics[width=0.24\linewidth]{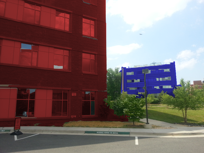}&
\includegraphics[width=0.24\linewidth]{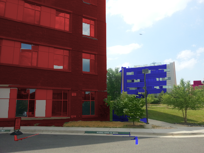}&
\includegraphics[width=0.24\linewidth]{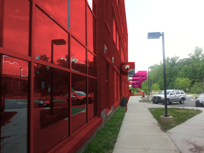}&
\includegraphics[width=0.24\linewidth]{labeling_img/20140522_150448_e}\\%
\includegraphics[width=0.24\linewidth]{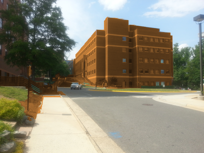}&
\includegraphics[width=0.24\linewidth]{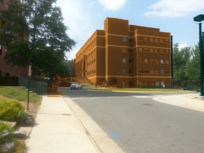}&
\includegraphics[width=0.24\linewidth]{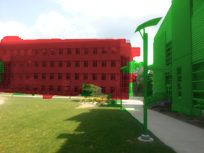}&
\includegraphics[width=0.24\linewidth]{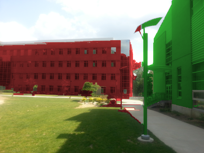}\\%
      \includegraphics[width=0.24\linewidth]{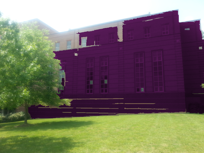}&
      \includegraphics[width=0.24\linewidth]{labeling_img/20140522_153047_e}&
      \includegraphics[width=0.24\linewidth]{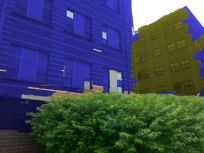}&
      \includegraphics[width=0.24\linewidth]{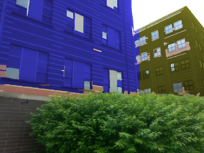}\\%
      \includegraphics[width=0.24\linewidth]{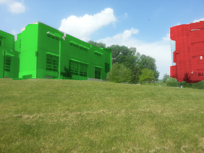}&
      \includegraphics[width=0.24\linewidth]{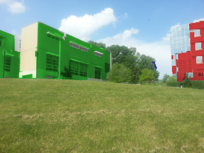}&
      \includegraphics[width=0.24\linewidth]{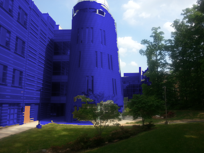}&
      \includegraphics[width=0.24\linewidth]{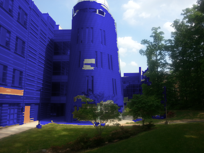}\\%
      (a) & (b) & (c) & (d)


 \end{tabular}
 \caption{Qualitative Comparison of the Proposed labeling method and Ground Truth. Columns (a) and (c) are the ground truth labeling of the reference view. Columns (b) and (d) show the building identity to pixel assignments using the most likely pose. Each building instance is shown with a different color. The output of our labeling shows that the orientation of the building sides are informative in the local neighberhood of the location of image acquisition. }
 \label{fig:trainingLabeling}
 \end{figure}
Consider that the map contains $s$ buildings  specified by their polygonal facades $\{ b_1, \hdots b_s \}$, where more than one building can be visible in the reference view.  We next  compute the per-pixel probability of a particular building, for each component of the semantic segmentation labelled as {\em building}.  Given the binary (building/no building) image and the inverse sensor model,  we can for each pose generate the expected horizontal extents of the buildings in the FOV. All the building image regions (pixels), whose horizontal extents 
overlap with the extents  projected from the map at a particular pose $g$ will be assigned label of the building $b_i$ with the probability of 
of that particular pose $p(\mathcal{I}|g)$ . 

\begin{figure}
 \includegraphics[width=0.47\linewidth]{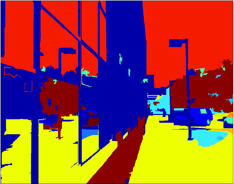}
  \includegraphics[width=0.49\linewidth]{labeling_img/20140522_150448_e}
 \caption{Semantic segmentation of an urban scene into four color-coded categories {\em building, trees, sky, vegetation and car}. The output of possibly noisy semantic segmentation is an input to our building identification and 3D reconstruction stages.} 
 \label{fig:SemanticSegmentation}
\end{figure}

Given an image $\mathcal{I}$ model and the position where image is taken we sample pose $g'$ such that the distance between $g'$ and the geo-tagged location of the image is less than a threshold. We can then follow two strategies: 1) maximum likelihood assignment: we chose the best camera pose $g^*$ and use $p_{pxl}(b_i \: | \mathcal{I},g^*)$ to find the assignment of building identities to pixels; 2) alternatively we can maintain all the probabilities and marginalize the camera poses. The building identity probability for each pixel is computed as:
\begin{equation}
\label{eq:sampling}
p_{pxl}(b_i | \mathcal{I}) \propto \Sigma_{g'} p(\mathcal{I}|g') \times p_{pxl}(b_i |,\mathcal{I},g')
\end{equation}
We assume the prior $p(g')$ around the geo-tagged location is uniform. We evaluate these two strategies in the experiments section. Figure~\ref{fig:trainingLabeling} illustrates the labeling of some of the images using the greedy approach.
\paragraph{Semantic Segmentation}
\label{sec:probForm}
Our approach for semantic labeling uses a single over-segmentation of the image where the superpixels are characterized with a variety of features including color, texture, location and perspective cues as in~\cite{Hoiem-IJCV07}. We also endow each superpixel region with a histogram of SIFT descriptors computed densely at each image location and quantized into 100 clusters. The entire feature vector is of 194 dimensions. 
Using boosting we learn classifiers for each of the five semantic categories {\em building, tree, road, sky, car} in a one vs. all fashion.
To train the classifier we use a  dataset of 320 images side views from StreetView sequence as in~\cite{QuanCVPR11}.  An example of the semantic segmentation of an image is shown in Figure ~\ref{fig:SemanticSegmentation}. 

\subsection{Building Retrieval}

The previous section described a pre-processing stage of the reference dataset of geo-tagged images. In this section, we want to use the result of the reference set with the identified and segmented buildings to infer the identity the buildings in the query image. We will use the inferred identity of the buildings in the query image along with the plane orientation information, to find the most likely location for the query image. We first retrieve top $k$ nearest view from the reference dataset using Semantically guided Bag-of-Words representation (SBoW). SBoW is a variation of BoW augmented with semantic information to enhance the matching process. For matching buildings, features that are from semantic classes other than buildings are removed from the BoW representation. The details of SBoW are described in \cite{Mousavian-ICRA-2015}. This strategy has been found to be effective against large changes in appearance due to seasonal changes as well as viewpoint variation. 

\begin{figure*}[t]
 \begin{tabular}{c@{\hspace{6mm}}c@{\hspace{6mm}}c@{\hspace{6mm}}c}
      \includegraphics[width=0.20\linewidth]{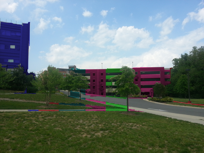}&
      \includegraphics[width=0.20\linewidth]{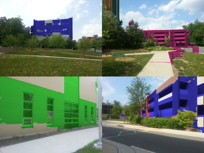}&
      \includegraphics[width=0.20\linewidth]{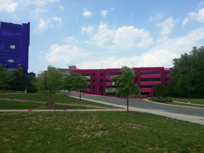}&
      \includegraphics[width=0.20\linewidth]{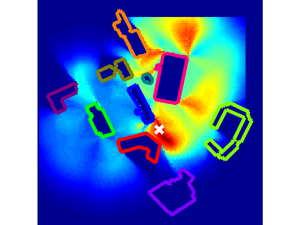}\\
      \includegraphics[width=0.20\linewidth]{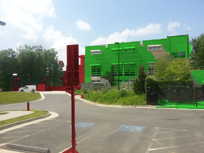}&
      \includegraphics[width=0.20\linewidth]{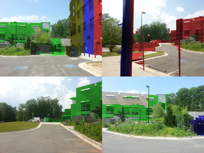}&
      \includegraphics[width=0.20\linewidth]{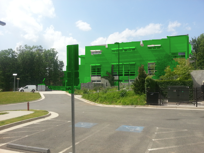}&
            \includegraphics[width=0.20\linewidth]{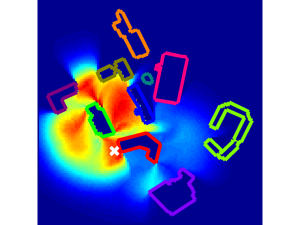}\\%
      \includegraphics[width=0.20\linewidth]{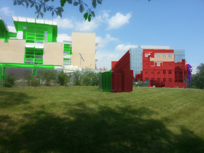}&
      \includegraphics[width=0.20\linewidth]{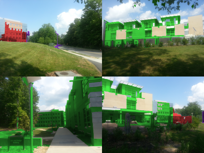}&
      \includegraphics[width=0.20\linewidth]{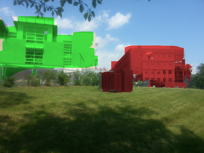}&   
            \includegraphics[width=0.20\linewidth]{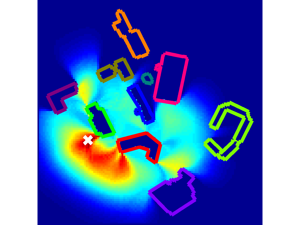}\\%
      \includegraphics[width=0.20\linewidth]{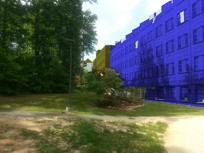}&
      \includegraphics[width=0.20\linewidth]{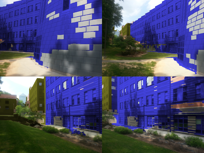}&
      \includegraphics[width=0.20\linewidth]{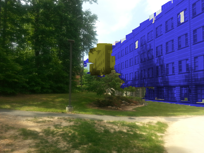}&  
            \includegraphics[width=0.20\linewidth]{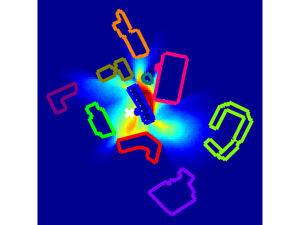}\\
(a) Query Image &
      (b) Retrieved Images &
      (c) Ground Truth & (d) Probability Map
      \\


 \end{tabular}
 \caption{Qualitative Evaluation of Identification of Query Image: (a) shows query image with inferred identity from the retrieval set. (b) shows the top four matches for the query image. Each of the retrieved set are colored with the maximum camera pose likelihood. (c) shows the ground truth labeling of the image. (d) is the probability map of the possible locations for the query image. White cross shows the GPS coordinate of the query image. Note that in the last row we missed the correct position but all the positions on the other side of the building got high probabilities.}
 \label{fig:testLabeling}
 \end{figure*}

\paragraph{Geolocation of the query view}
Authors in \cite{Mousavian-ICRA-2015} proposed a method to identify the buildings and estimate their horizontal extent in the query image. The horizontal extent of a buildings can be spanned by multiple visible facades. The goal of this matching stage is to compute the probability of building identity using the SIFT features matched with the reference views.  SIFT features in the reference image have probabilities of coming from different buildings and when matched as the inlier features,  the building identity probabilities are transferred into the query image. 
This information is used to find horizontal extent of the buildings in addition to their identity. 
\noindent
Now the problem is similar to what we had for camera pose likelihood estimate for the reference set. 
The only difference is that now we have additional evidence about the building identities for both of the $z_i^k$ and $\widehat{z_j}^k$ coming from 
SIFT matching between the query view and the reference set. This building identity will also be used to determine the approximate geo-location for more detailed pose sampling. 
In order to geo-locate  the query image we sample the poses $g$ around the identified buildings and for each sampled camera pose, we compute the likelihood in Eq \ref{eq:LL} with considering the identity of $z_i^k$ and $\widehat{z_j}^k$. The equivalent of $\mathds{1}(z_i^{k} \cap \widehat{z_j}^k \neq \emptyset)$ with considering the identity is the set of columns $k$ where there is at least one plane with the same identity in both of the $z_i^k$ and $\widehat{z_j}^k$. The rest of the terms remains the same. Figure~\ref{fig:testLabeling} shows the qualitative results for building identification of the query images. Since the building identities are propagated from all the retrieved images, discrepancies in some of the retrieved images will be resolved by others. The first two rows of  Figure~\ref{fig:testLabeling} are examples of  situations where the retrieved images contain incorrect building identities. 

\section{Experiments}
We chose a subset of 270 images in the dataset of \cite{Mousavian-ICRA-2015}. The images are taken from 11 buildings at a University campus. The instances of buildings are segmented for each image using semantic segmentation which we will use as the ground truth segmentation. 
This segmentation is not perfect but they covers most of the buildings. The dataset has sparse set of views and it also contains images where the building is not in the center of FOV. 
For retrieving the images, we closely follow the approach of \cite{Mousavian-ICRA-2015} where they use SBoW to retrieve images with most overlap with the query image.
\paragraph{Reference Set Building Identification} 
In this section we report the accuracy of our building identification strategy given a map for the set of reference views.
Recall that the orientation of the reference images is unknown. Given the correct orientation and location estimate the buildings projected
to the image from the map should have the perfect overlap with the buildings detected in the image. 
However in order to compensate for GPS tags errors for each reference image, we divide the neighborhood of the reference image location in to $3 \times 3$ grid centered at tagged location where each point is 10 meter away from its neighbors. For each of the sampled locations, we evaluate  the pose likelihood equation Eq~\ref{eq:LL}  and for each location we check 120 different orientations. Given sampled poses,  where each will yield a projection of the 
buildings from the map to an image, Eq~\ref{eq:sampling} is used to compute per-pixel building identities. We then compute the ratio of the identified building pixels
to the ground truth pixels identified as buildings in the image. The final accuracy will depend on whether we will compute this 
ratio only for a single most likely pose or marginalize over the neighboring poses.  Since the proposed method does not assign any identity to the pixels which are not in the regions classified as building, we compared only the pixels which are labeled as buildings. 
As Table \ref{tab:refIdentify} shows the marginalized assignment performs better. This is due to the fact that in some locations, there are multiple likely hypothesis. In such situations, when we are choosing the best hypothesis using the greedy algorithm we might lose another highly likely hypothesis which were correct indeed. 

\paragraph{Geolocation of Query View} In order to geo-locate the query views, similar images are retrieved from the reference set.  Using the retrieval set, the identity of buildings in the query image is inferred. After identifying the buildings in the query image, we discretize the location into a grid of locations where adjacent locations are 10 meters from each other. For each location, 120 different orientations are evaluated and identity-aware version of Eq~\ref{eq:sampling} is computed. Due to the inherent ambiguities of the localization, the pose likelihood does not have a single maximum for some of the images (See Figure~\ref{fig:FOV}). Therefore, we compute a pose probability map for each query image. We divide the evaluation of the query image into two criteria: 1) Accuracy of building identifications in the query image; 2) Localization accuracy of the query images. 

\noindent For evaluation of building identification accuracy, we compared greedy pose selection and marginalized sampling against the supervised situation where we use the ground truth labeled identity of the buildings. We took the pixel level identification which was available in the dataset and compared the accuracy of pixel identification in Table~\ref{tab:qIdentity}. As Table~\ref{tab:qIdentity} shows our unsupervised building identification, where the building identities in the reference set were determined  using our pose estimation strategy given a map, gives comparable performance with respect to the supervised method that requires pixel level labeling for identity of building instances. 

\begin{figure}
\includegraphics[width=0.40\textwidth]{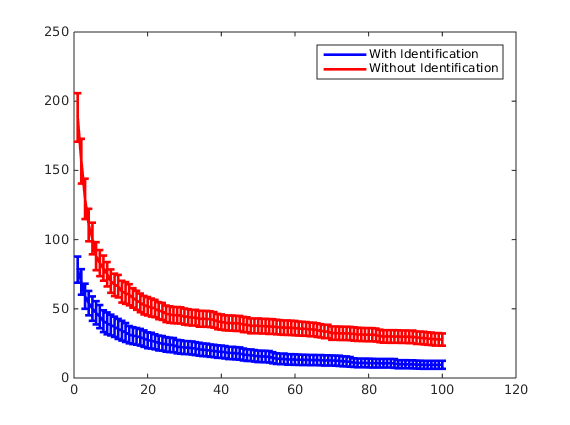}
\caption{Comparison of average localization error with 95\% confidence interval (in meter) using building identification and without building identification using the $N$ most likely locations (x-axis). }
\label{fig:localizerion_err}
\end{figure}

\noindent One of the main differences between our approach and approaches of \cite{Cham-CVPR-2010} and \cite{David-IROS-2011} is that they only use geometry signature of the $360^\circ$ FOV image to localize the query image. We proposed that geometric signature is not enough for localization of query images taken by limited FOV camera. As a result, we compare the accuracy of our method with building identification versus without building identification.  As mentioned before, the location of some of the query images can be ambiguous. In order to quantitatively measure the localization error, we need to define the error. For each computed probability map, we find the top $N$ locations with highest probability. The localization error for each query image is the minimum distance between the ground truth location and top $N$ locations with the highest probability. Figure~\ref{fig:localizerion_err} shows the plot of average error using the top $N$ likely positions. As it is shown in Figure~\ref{fig:localizerion_err}, localization using geometric signature and building identification is statistically better than localization using only geometric signature.

\begin{table}[t]
\centering
\caption{Quantitative comparison of Accuracy between Greedy vs. Marginalized Assignment }
\label{tab:refIdentify}
\begin{tabular}{|c | c|}
\hline
Greedy Assignment & Marginalized Assignment \\
\hline
0.7466 & {\bf 0.7627}\\
\hline
\end{tabular}
\end{table}

\begin{table}[t]
\centering
\caption{Pixel accuracy of ground truth identification(supervised) vs. unsupervised assignment}
\begin{tabular}{| c | c | c | c | }
\hline
$k$ & Ground Truth & Greedy  & Marginalized\\
\hline
1 & 0.7543 & 0.7365 & 0.7363\\
4 & 0.8113 & 0.8016 & 0.7915\\
8 & 0.8254 & 0.8144 & 0.8177\\
\hline
\end{tabular}
\label{tab:qIdentity}
\end{table}

\section{Conclusions}

We have demonstrated a novel approach for semantic geolocation using a map. The proposed method extends the types of environments where images can be geolocated to settings where sparse set of reference views and the map of the area are available. 
Using the map constraints together with techniques for semantic segmentation and reconstruction from a single view we can 
estimate the camera orientation for  the reference view and  automatically label the identities of individual buildings in images.
The final pose likelihood map is computed by combining different sources of evidence coming from appearance based matching, 
the meta information associated with the map and a single view reconstruction. 
This enhanced labeling then improves the BoW retrieval strategies in case the novel query views have a small visual overlap
and enables geolocation of novel view combining different sources of evidence.

 \linespread{0.78}
{\footnotesize
\noindent {\bf Acknowledgements}
Supported by the Intelligence Advanced Research Projects Activity (IARPA) via Air Force Research Laboratory, contract FA8650-12-C-7212. The U.S. Government is authorized to reproduce and distribute reprints for Governmental purposes notwithstanding any copyright annotation thereon. Disclaimer: The views and conclusions contained herein are those of the authors and should not be interpreted as necessarily representing the official policies or endorsements, either expressed or implied, of IARPA, AFRL, or the U.S. Government.
}
 
\bibliographystyle{plain}%
\bibliography{root}

\end{document}